\documentclass[conference]{IEEEtran}
\IEEEoverridecommandlockouts
\usepackage{cite}
\usepackage{amsmath,amssymb,amsfonts}
\usepackage{algorithm, algorithmic}
\usepackage{graphicx}
\usepackage{subcaption}
\usepackage{caption}
\usepackage{textcomp}
\usepackage{float}
\usepackage{xcolor}
\usepackage{amsmath}
\usepackage{bm}
\usepackage{url}
\usepackage{booktabs}

\newcommand{\bftab}{\fontseries{b}\selectfont}

\def\BibTeX{{\rm B\kern-.05em{\sc i\kern-.025em b}\kern-.08em
    T\kern-.1667em\lower.7ex\hbox{E}\kern-.125emX}}
\begin{document}

\title{Comprehensive Semantic Segmentation on High Resolution UAV Imagery for Natural Disaster Damage Assessment\\
{
}
}
\author{
\IEEEauthorblockN{Maryam Rahnemoonfar, Tashnim Chowdhury}
\IEEEauthorblockA{\textit{Computer Vision and Remote Sensing Laboratory} \\
\textit{University of Maryland Baltimore County}\\
Baltimore, USA \\
maryam, tchowdh1@umbc.edu}

\and
\IEEEauthorblockN{Robin Murphy, Odair Fernandes}
\IEEEauthorblockA{\textit{Department of Computer Science and Engineering} \\
\textit{Texas A\&M University}\\
College Station, USA\\
murphy@cse.tamu.edu, odair\_fernandes@tamu.edu}
} 

\maketitle

\begin{abstract}
In this paper, we present a large-scale hurricane Michael dataset for visual perception in disaster scenarios, and analyze state-of-the-art deep neural network models for semantic segmentation. The dataset consists of around 2000 high-resolution aerial images, with annotated ground-truth data for semantic segmentation. We discuss the challenges of the dataset and train the state-of-the-art methods on this dataset to evaluate how well these methods can recognize the disaster situations. Finally, we discuss challenges for future research.
\end{abstract}


\section{Introduction}

In recent time, the world has seen numerous natural disasters which have brought both personal injury and economic loss to several countries all over the world including USA. In 2019 USA has inflicted with 14 natural disasters which have cost around 45.4 billions dollars and till May 2020, 10 natural disasters have been reported with total economic loss of approximately 17.6 billions dollars \cite{noaa}. A major step to both save valuable human lives and reduce financial loss is an accurate assessment of the damage inflicted by these events. A correct estimation of the damage helps to plan efficiently and allows the rescue team to allocate their efforts and aid properly. However, current available damage assessment procedures are manual which include field supervision and reports which is very difficult to obtain and even sometimes impossible due to inaccessible heavily affected areas.

With the advent of new sophisticated technologies capturing events of natural disaster has improved. Currently aerial and satellite imageries \cite{chen2018benchmark, gupta2019creating} have been used for proper assessment of these disaster events \cite{chen2018benchmark, gupta2019creating}. The rescue team can use UAV (Unmanned Aerial Vehicle) to capture images of damaged properties and the whole affected area. Compared to satellite imagery, UAV imagery provides better resolution which helps to understand the detailed damage level of the captured area. Recently researchers have been working on evaluating the damages caused by different natural disasters using DCNN (Deep Convolutional Neural Network) \cite{lopez2017river, doshi2018satellite, rahnemoonfar2018flooded, rudner2019multi3net, gupta2020rescuenet, gupta2020deep, zhu2020msnet}. These research works mostly involve in detecting different damaged buildings and roads, and sometimes only flooded areas. Few researchers have worked on detecting and classifying the damage levels of the buildings. During natural disasters debris is also an important part for the estimation of damage level of an area. But no works have been found which attempts to detect debris along with buildings, roads, and flood water. 

Complete image understanding is one of the primary research areas nowadays due to its primal part in numerous state-of-art applications such as autonomous driving \cite{ess2009segmentation, geiger2012we, cordts2016cityscapes}, human behavior analysis \cite{cao2017realtime}, face recognition \cite{yang2016multi}, computational photography \cite{yoon2015learning}, and image search engines \cite{wan2014deep}. Semantic segmentation is a core part of image understanding. Semanctic segementation is the task of assigning semantic labels to every pixels of an image. Besides traditional approach to tackle computer vision problems like semantic segmentation, deep convolutional neural networks  \cite{sermanet2013overfeat, long2015fully-fcn} have achieved several revolutionary achievements in answering complex image understanding issues. Although many advanced segmentation works have been proposed in the recent years for popular urban datasets like Cityscapes \cite{cordts2016cityscapes} and PASCAL VOC \cite{mottaghi_cvpr14-pascal} dataset, very few methods \cite{doshi2018satellite, rudner2019multi3net, gupta2020rescuenet, zhu2020msnet, rahnemoonfar2018flooded} have been proposed and applied on natural disaster datasets. 

Three major issues with the application of deep convolutional neural networks for semantic segmentation are: the lose of information or decrement of signal resolution, objects with different sizes, and spatial invariance, The first problem is caused due to downsampling and pooling operation in the neural networks. The second issue is related to the presence of objects of multiple scales. And the third one, spatial invariance, refers to the gain of low spatial information at the higher layers of convolutional networks.

\begin{figure*}[t]
	\begin{center}
		\includegraphics[width=\linewidth]{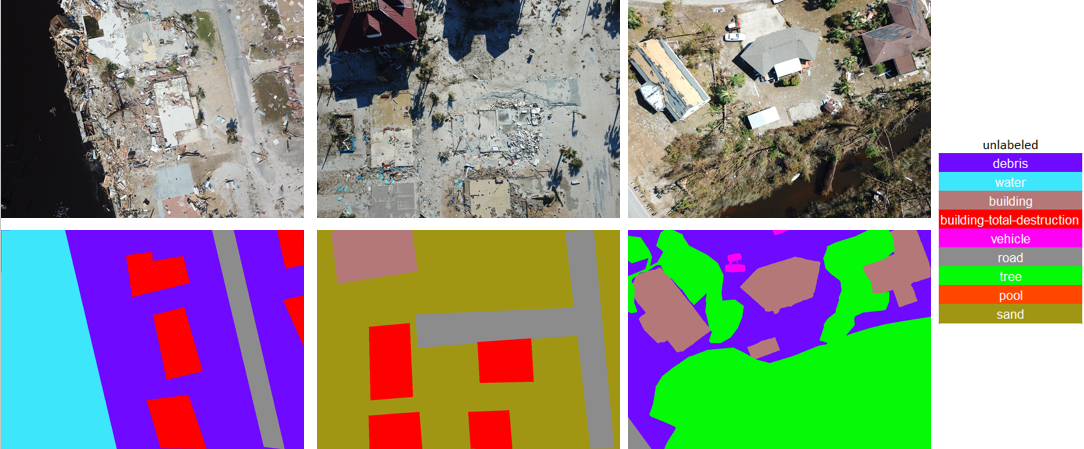}
	\end{center}
	\caption{Illustration of complex scenes of HRUD dataset. First row shows original image and the second row shows the corresponding annotations.}
	\label{fig:hrud-intro-exm-with-label}
\end{figure*}

In this paper, we address both semantic segmentation issues and issues related to natural disaster dataset. At first, we propose a new high resolution UAV dataset named HRUD (High Resolution UAV Dataset) where resolution of the captured images are 3000 $\times$ 4000. Data are collected after Hurricane Michael. We analyze and evaluate performance of popular semantic segmentation methods named DeepLabv3+ \cite{chen2018encoder-deeplabv3+}, PSPNet \cite{zhao2017pyramid-pspnet}, and ENet \cite{paszke2016enet} on these images. All these methods attempt to address the three segmentation problems mentioned earlier. We also discuss the challenges of the dataset and scrutinize the performances of the three methods on it. Compared to other works on disaster dataset, we comprehensively perform semantic segmentation on all of the components present in the images. Figure \ref{fig:hrud-intro-exm-with-label} shows sample images and respective colored annotated masks from HRUD dataset. From the best of our knowledge HRUD is the largest high resolution dataset from a single natural disaster and also this paper presents the first complete scene segmentation on a natural disaster dataset.


The reminder of this paper is organized as follows: it begins with highlighting the current advances on natural disaster damage assessment and semantic segmentation in section \ref{related-works}. Next section \ref{dataset} describes the HRUD dataset including its annotation, and challenges. The next section \ref{network-models} shows a brief review on three semantic segmentation methods.  Section \ref{experiment} explains the experimental setup for each network models and section \ref{ablation} shows performance of the models on different experimental setup. Finally section \ref{discussion} summarizes the results while conclusion and future works are mentioned in section \ref{conclusion}.

\section{Related Works} \label{related-works}

\subsection{Natural Disaster Damage Assessment Dataset}

Existing natural disastar damage assessment datasets can be categorized into two types: ground-level images \cite{nguyen2017damage}, and satellite imagery \cite{chen2018benchmark, gupta2019creating}. The ground-level images were mostly collected from social media \cite{nguyen2017damage}. The issues related to those datasets are scarcity of images along with lack of geo location tags. The second type of dataset is satellite imagery \cite{chen2018benchmark, gupta2019creating} which is collected using remote sensing equipment. One of the major issues of this type of dataset is the lack of detailed information about the damaged area since vertical viewpoint has limited access to damage information which might be easier to obtain from horizontal viewpoint. This paper introduces the largest high resolution natural disaster dataset collected from a single disaster using UAVs.

\subsection{Natural Disaster Damage Assessment Methodologies}
Recently several research works have been proposed addressing the damage assessment from natural disaster datasets. In \cite{lopez2017river} a dataset with 300 images are used for river segmentation to monitor flood water where three already proposed methods \cite{long2015fully-fcn, jegou2017one-tiramisu, isola2017image-pix2pix} are analyzed and evaluated. Authors in \cite{doshi2018satellite} perform semantic segmentation on satellite images to detect changes in the structure of various man-made features, and thus detect areas of maximul impact due to natural disaster. UAV images have been used for flood area detection by Rahnemoonfar et. al. in \cite{rahnemoonfar2018flooded} while presenting a densly connected recurrent neural network. Rudner et. al. present a novel approach named Multi3Net in \cite{rudner2019multi3net} for rapid segmentation of flooded buildings by fusing multiresolution, multisensor, and multitemporal satellite imagery. RescueNet is proposed by Gupta et. al. in \cite{gupta2020rescuenet} for joint building segmentation. In \cite{gupta2020deep} authors analyzed popular segmentation models on aerial image dataset by performing semantic segmentation on buildings and roads. Zhu et. al. propose a multilevel instance segmentation named MSNet in \cite{zhu2020msnet} on aerial videos to assess building damage after natural disaster.

In this paper, we evaluate three state-of-art segmentation network models, ENet \cite{paszke2016enet}, DeepLabv3+ \cite{chen2018encoder-deeplabv3+}, and PSPNet \cite{zhao2017pyramid-pspnet}, on our newly proposed HRUD dataset. Our approach is similar to \cite{lopez2017river, gupta2020deep} from the perspective of evaluating already proposed network models, but the prime difference from all of the works mentioned before is that our work performs a comprehensive segmentation on all of components of the images in the dataset. Segmentation performed on the HRUD dataset is not limited on only to detect roads and damaged buildings, but all components present in the images including debris, water, building, road, tree, vehicle, pool, and sand. We also segment the buildings based on damage levels in a separate experiment.

\subsection{Semantic Segmentation}
Semantic segmentation is one of the core tasks of vision problem. Recently with the application of deep learning more advanced methods have been proposed. The fully convolutional networks (FCN) \cite{long2015fully-fcn} is one of the pioneer methods in semantic segmentation which replaces the last fully connected layer of deep convolutional neural network by convolutional layers. FCN uses upsampling and concatenation of updates from inter-mediate feature maps to address spatial invariance problem. To extract contextual information to resolve decrease of signal resolution, several methods have been proposed. Some methods have adopted multi-scale inputs as a form of pyramid pooling while other methods have implemented probabilistic graphical methods, Conditional Random Fields (CRF). Another type of deep convolutional neural network is called encoder-decoder type architecture.

Here pyramid pooling based methods and encoder-decoder based methods are briefly discussed. Models, such as PSPNet \cite{zhao2017pyramid-pspnet} or DeepLab \cite{chen2017rethinking-deeplabv, chen2017deeplab}, perform spatial pyramid pooling \cite{grauman2005pyramid, lazebnik2006beyond} at several grid scales (including image- level pooling \cite{liu2015parsenet}) or apply several parallel atrous convolution with different rates (called Atrous Spatial Pyramid Pooling, or ASPP). These models have shown promising results on several segmentation benchmarks by exploiting the multi-scale information.

The encoder-decoder networks have been successfully applied to many computer vision tasks, including human pose estimation \cite{newell2016stacked}, object detection \cite{lin2017feature, shrivastava2016beyond, fu2017dssd}, and semantic segmentation \cite{long2015fully-fcn, ronneberger2015u-unet, badrinarayanan2017segnet, noh2015learning, lin2017refinenet, pohlen2017full, peng2017large, amirul2017gated, wojna2017devil, fu2019stacked, zhang2018exfuse}. Typically, the encoder-decoder networks are consist of a encoder and a decoder module. The encoder part transforms the feature maps into smaller ones which hold higher semantic information. On the other hand. a decoder translates the spatial information. Some notable works using this architecture are \cite{ronneberger2015u-unet , chen2017rethinking-deeplabv}.

These state-of-art semantic segmentation networks have been mainly applied on ground based imagery \cite{cordts2016cityscapes, mottaghi_cvpr14-pascal}. Although few research works \cite{lopez2017river, gupta2020deep} have applied few popular network models on aerial imagery, these works do not include a complete semantic segmentation of these images. In contrast to other research works, we apply three state-of-art semantic segmentation network models on our proposed HRUD dataset for a complete scene segmentation. We adopt one encoder-decoder based network named ENet \cite{paszke2016enet}, one pyramid pooling module based network PSPNet \cite{zhao2017pyramid-pspnet}, and the last network model DeepLabv3+ \cite{chen2018encoder-deeplabv3+} employs both encoder-decoder and pyramid pooling module.

\begin{figure*}[t]
	\begin{center}
		\includegraphics[width=\linewidth]{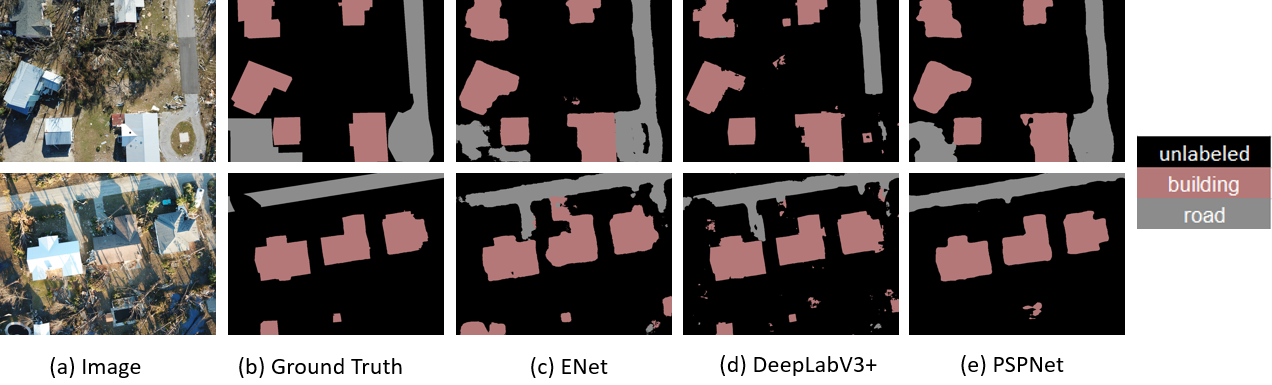}
	\end{center}
	\caption{Visual comparison on HRUD test set for Experiment 1.}
	\label{fig:vis-compare-models-exp-1}
\end{figure*}

\begin{figure*}[t]
	\begin{center}
		\includegraphics[width=\linewidth]{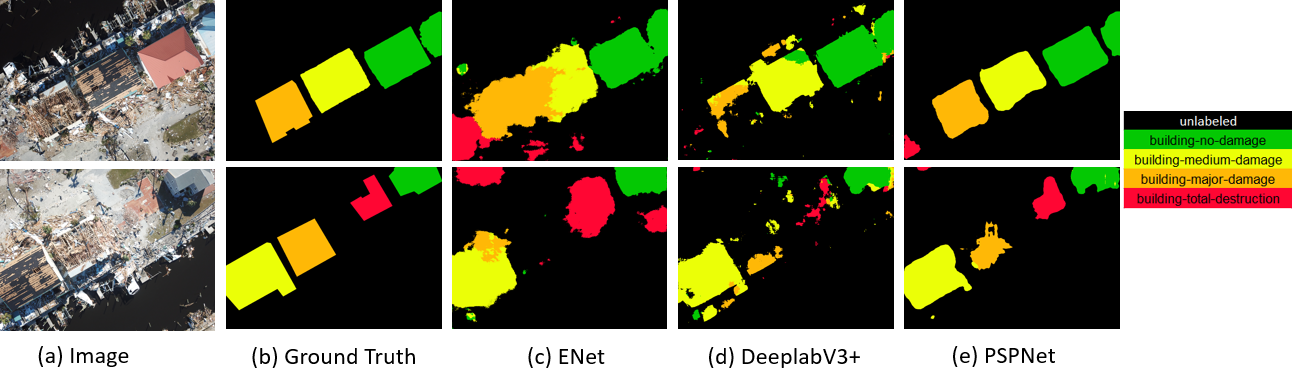}
	\end{center}
	\caption{Visual comparison on HRUD test set for Experiment 2.}
	\label{fig:vis-compare-models-exp-2}
\end{figure*}

\begin{figure*}[t]
	\begin{center}
		\includegraphics[width=\linewidth]{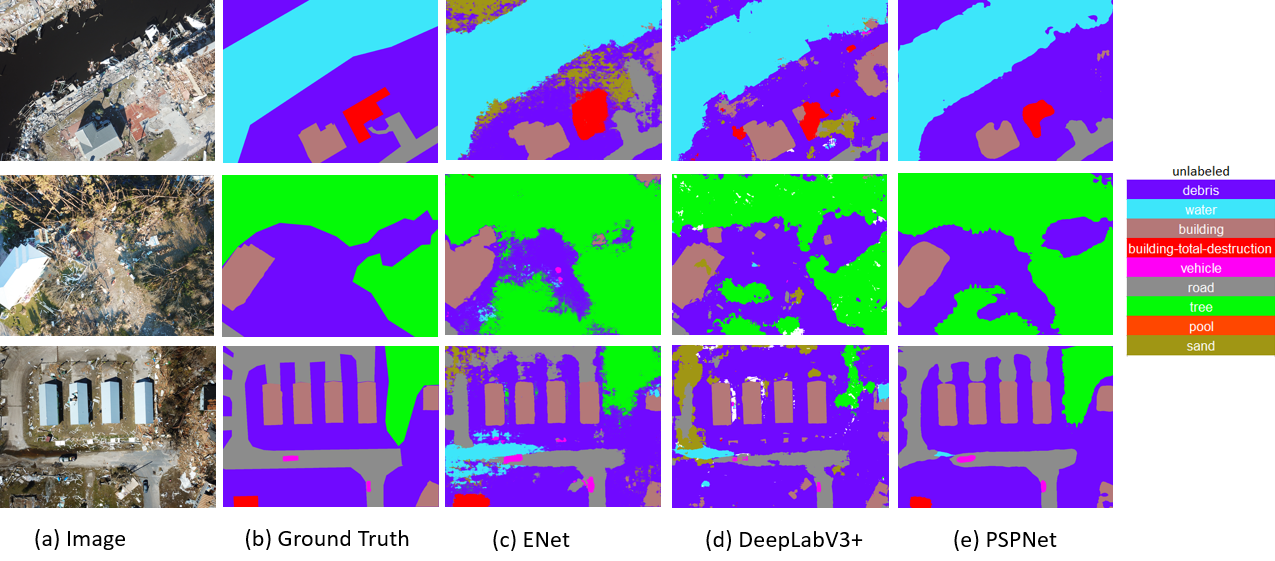}
	\end{center}
	\caption{Visual comparison on HRUD validation set for Experiment 3.}
	\label{fig:vis-compare-models-all-cls-val}
\end{figure*}

\begin{figure*}[t]
	\begin{center}
		\includegraphics[width=\linewidth]{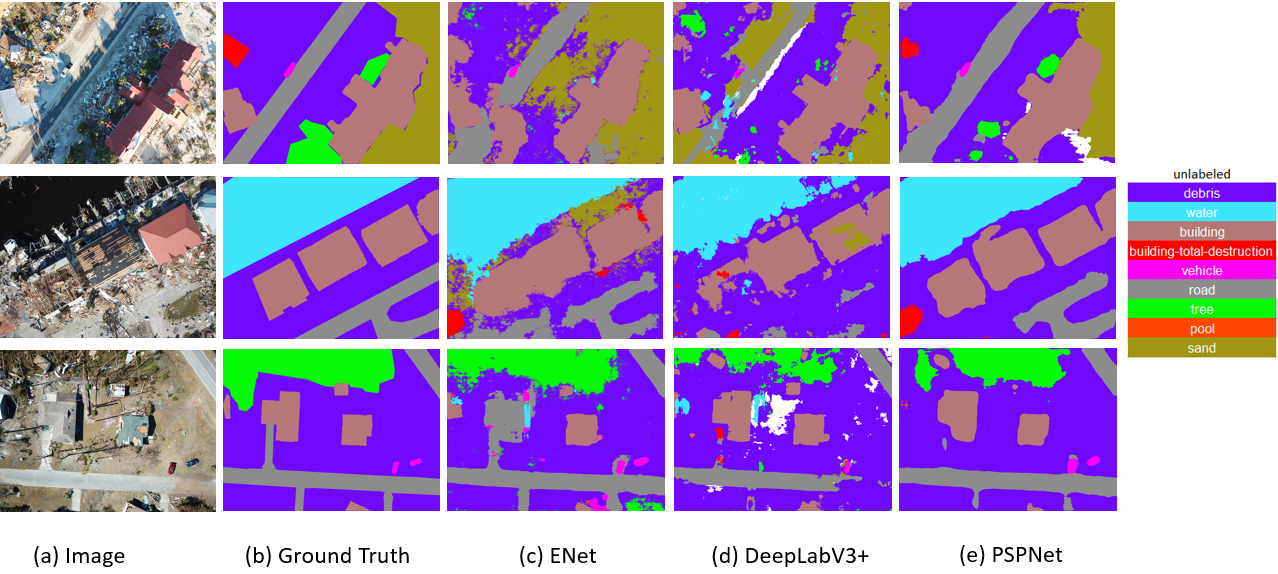}
	\end{center}
	\caption{Visual comparison on HRUD test set for Experiment 3.}
	\label{fig:vis-compare-models-all-cls}
\end{figure*}

\section{Dataset}\label{dataset}

\subsection{Data Annotation} \label{section:data-annotation}
We annotate the dataset for semantic segmentation. Annotations are performed on the V7 Darwin platform \cite{V7Darwin}. The main objective of the semantic segmentation annotation is to annotate all the objects present in the images including debris, water, building, vehicle, road, tree, pool, and sand. The motivation behind complete semantic segmentation is to annotate all the damaged objects in an image, since it is important to identify all the damaged objects in an image to evaluate the actual damage done by a natural disaster. In total 1973 images have been annotated for semantic segmentation. Most of the images include debris and buildings. The buildings include both residential and non-residential structures. The buildings are annotated into four classes based on their damage levels. The classes are no damage, medium damage, major damage, and total destruction. Number of annotated polygons of buildings of these four classes are shown in Table \ref{table:data-stat}.

\begin{table}[!htp]
	\centering
	\caption{Number of polygons of different buildings based on their damage levels.}\label{table:data-stat}
	\begin{tabular}{l c}
	\hline
		Damage Level & Number of Polygons\\\hline
		\hline
		No Damage & 1738 \\
		Medium Damage & 1424\\
		Major Damage & 686\\
		Total Destruction & 1255\\ \hline
	\end{tabular}
\end{table}

If a building is unharmed during the disaster event, then this building is classified as Building-No-Damage. If some parts of a building are damaged, but these parts can be covered with a blue tarp, then the building is classified as Building-Medium-Damage. On the other hand, if the roof of a building is damaged such a way that it has sustained significant structural damage and requires extensive repairs,  then the building is denoted as Building-Major-Damage. And when complete failure of two or more major structural components - e.g. collapse of basement walls, foundation, load-bearing walls, or roof is observed, then the building is classified as Building-Total-Destruction.
 
 Compared to other disaster dataset \cite{chen2018benchmark, gupta2019creating}, we annotate all of the components in the dataset which includes debris, water, buildings, vehicle, road, tree, pool, and sand. This allows us to perform complete segmentation of the images. When all other disaster datasets can only be used for binary segmentation \cite{chen2018benchmark} (flooded or non-flooded area, damaged building or undamaged building) or only to detect different damage levels of buildings \cite{gupta2019creating}, our dataset provides an opportunity for the researchers to perform comprehensive segmentation on all of the damaged and undamaged components of the images. 

\subsection{Dataset Challenges} \label{section:dataset-challenges}

There are several challenges posed by this dataset. One of the challenges is difficulty in differentiating between damage levels of different building instances. Since UAV image only include top view of a building, it is very difficult to estimate how much damage is done on that building. Horizontal view brings most information about a building's current damage condition but the UAV images provide only top views. We would like to investigate if the deep neural network models will have any difficulty in evaluating the damage levels of the buildings. Similarly another prime challenge is identifying the roads since several times roads are covered with debris, and sometimes only a small portion of a whole road is visible. Since debris are scattered all over the place, it will also be difficult to segment the sand and water. A lot of debris are fallen on the water and sand. Moreover the textures of the debris and sand are very similar and tough to separate from each other during semantic segmentation which might bring down the performance of the network models. During the disaster event a lot of trees are damaged and they are scattered along with the debris. Differentiating between undamaged trees and damaged trees is also a challenging task for the network models. If a tree is damaged due to the hurricane, we consider it as part of the debris. Due to the immense destruction on both natural and man-made structures, the dataset contains very complex and complicated structures. For example, the debris from buildings and vehicles are combined together in many images. Many houses are destroyed and it is very difficult to identify totally destroyed buildings from debris. Specially since this is an aerial image, determining the actual destruction level of a building is very difficult. Finally segmenting vehicles is also challenging since sometimes vehicles can be found in the middle of the debris. If vehicles are mixed with debris, the vehicles are considered as part of debris. Therefore the models have to understand the scenario where vehicles should be segmented as vehicles and where it should be segmented as part of debris.

Figure \ref{fig:hrud-intro-exm-with-label} shows few images and respective colored annotated masks from HRUD dataset. The first image indicates a mixture of debris and building-total-destruction. Although in the first image the road is visible, it is very difficult to distinguish between debris and building-total-destruction instances. The second image introduces another challenge of segmenting roads along with already discussed building-total-destruction. The roads are almost invisible due to sand and debris. The third image presents another challenge for the segmentation models where we can see two different types of building damages, two buildings on the right with no damage and the building with medium damage on the left.

\section{Methods} \label{network-models}
We employ three state-of-art semantic segmentation models named ENet \cite{paszke2016enet}, PSPNet \cite{zhao2017pyramid-pspnet}, and DeepLabv3+ \cite{chen2018encoder-deeplabv3+} to evaluate their performance on our proposed HRUD dataset. We choose three different network models from two types of neural networks, spatial pyramid pooling module and encoder-decoder structure. Among the selected three models, ENet \cite{paszke2016enet} adopt encoder-decoder structure based model while PSPNet \cite{zhao2017pyramid-pspnet} employs pyramid pooling module based model. DeepLabv3+ \cite{chen2018encoder-deeplabv3+} is a fusion between these two types of network designs.

\subsection{ENet}
ENet is a state-of-art encoder-decoder structure module based segmentation model which performs excellent compared to concurrent encoder-decoder based segmentation models \cite{badrinarayanan2017segnet, ronneberger2015u-unet} on popular datasets like Cityscapes \cite{cordts2016cityscapes} and CamVid \cite{brostow2008segmentation-camvid}. In ENet \cite{paszke2016enet} authors have proposed a compact encoder-decoder architecture which attempts to employ different optimization details in the architecture to address current research issues. ResNet\cite{he2016deep-resnet} has been adopted as backbone architecture which consists of a single main branch and later it has been added with extensions consists of convolutional filters by element-wise addition. The architecture has consists of two modules, inital and bottleneck module. ENet initial block performs both convolution and max pooling on input and concatenate the results.  The Bottleneck layer, adopted from ResNet, includes three convolutional layers where Batch Normalization and PReLU is placed between all convolutions. PReLUs instead of ReLUs, an additional parameter to learn in each feature map, is used with an objective to learn negative slope of non-linearities. Bottleneck block is used for both upsampling and downsampling based on decoder and encoder. A max pooling layer is added to the bottleneck layer when this layer is used for downsampling. The ENet encoder-decoder architecture is shown in Table \ref{enet-table}.

\begin{table}[htp]
	\centering
	\caption{ENet architecture. Output sizes are given
for an example input of 512 $\times$ 512.}\label{enet-table}
	\begin{tabular}{l c l}
		Name   & Type & Output  Size\\\hline
		\hline
		initial &  & 16 $\times$ 256 $\times$ 256\\
		\hline
		bottleneck1.0 & downsampling & 64 $\times$ 128 $\times$ 128 \\
		4 $\times$ bottleneck1.x & & 64 $\times$ 128 $\times$ 128\\ \hline
		bottleneck2.0 & downsampling & 128 $\times$ 64 $\times$ 64 \\
		bottleneck2.1 &  & 128 $\times$ 64 $\times$ 64 \\
		bottleneck2.2 & dilated 2 & 128 $\times$ 64 $\times$ 64 \\
		bottleneck2.3 & asymmetric 5 & 128 $\times$ 64 $\times$ 64 \\
		bottleneck2.4 & dilated 4 & 128 $\times$ 64 $\times$ 64 \\
		bottleneck2.5 &  & 128 $\times$ 64 $\times$ 64 \\
		bottleneck2.6 & dilated 8 & 128 $\times$ 64 $\times$ 64 \\
		bottleneck2.7 & asymmetric 5 & 128 $\times$ 64 $\times$ 64 \\
		bottleneck2.8 & dilated 16 & 128 $\times$ 64 $\times$ 64 \\ \hline
		bottleneck3.1 &  & 128 $\times$ 64 $\times$ 64 \\
		bottleneck3.2 & dilated 2 & 128 $\times$ 64 $\times$ 64 \\
		bottleneck3.3 & asymmetric 5 & 128 $\times$ 64 $\times$ 64 \\
		bottleneck3.4 & dilated 4 & 128 $\times$ 64 $\times$ 64 \\
		bottleneck3.5 &  & 128 $\times$ 64 $\times$ 64 \\
		bottleneck3.6 & dilated 8 & 128 $\times$ 64 $\times$ 64 \\
		bottleneck3.7 & asymmetric 5 & 128 $\times$ 64 $\times$ 64 \\
		bottleneck3.8 & dilated 16 & 128 $\times$ 64 $\times$ 64 \\ \hline
		bottleneck4.0 & upsampling & 64 $\times$ 128 $\times$ 128 \\
		bottleneck4.1 &  & 64 $\times$ 128 $\times$ 128 \\
		bottleneck4.2 &  & 64 $\times$ 128 $\times$ 128 \\ \hline
		bottleneck5.1 & upsampling & 16 $\times$ 256 $\times$ 256 \\
		bottleneck5.2 &  & 16 $\times$ 256 $\times$ 256 \\ \hline
		fullconv & & C $\times$ 512 $\times$ 512 \\ \hline
	\end{tabular}
\end{table}

Although downsampling is used in ENet as a part of encoder architecture, it has followed SegNet \cite{badrinarayanan2017segnet} approach which saves the indices of the elements chosen in max pooling in order to reduce the memory requirements. SegNet is a encoder-decoder based model which has almost symmetrical encoder and decoder structure similar to U-Net \cite{ronneberger2015u-unet}. But the encoder-decoder architecture of ENet is different from SegNet, since in ENet authors have proposed large encoder and relatively smaller decoder. The reasoning behind this design is that decoder only upsamples the output of the encoder to fine-tune the details, therefore it is not required to be large like encoder which works on smaller resolution data for data processing and filtering. Dilated convolution has been used to wider the context by having the wider receptive field since the network already hurts the accuracy of the network by downsampling the feature maps. Spatial Dropout has been used as regularizer at the end of convolutional branches.

\subsection{PSPNet}
The motivation behind adopting PSPNet \cite{zhao2017pyramid-pspnet} is to evaluate the state-of-art pyramid pooling based segmentation model on our HRUD dataset. PSPNet \cite{zhao2017pyramid-pspnet} shows pioneering performance on popular datasets \cite{cordts2016cityscapes, zhou2017scene-ade20k}. Since our dataset is a natural disaster dataset, there are a lot of debris and damaged structures scattered all around the area. Global context prior would perform tremendously in segmenting individual damaged and undamaged components in such aerial imagery which ultimately motivates us to choose PSPNet \cite{zhao2017pyramid-pspnet} as one the three methods for evaluating on HRUD. 

With an objective to increase the size of empirical receptive field of convolutional neural networks, global contextual prior, Zhao et. al. has proposed PSPNet (pyramid scene parsing network) \cite{zhao2017pyramid-pspnet}. Global average pooling is presented as global contextual prior in \cite{liu2015parsenet} for semantic segmentation. Extraction of global context information along with local context information from different sub-regions helps to distinct among objects of different categories in complex datasets. This idea is proposed in \cite{he2015spatial-sppnet} for classification task. PSPNet further improves the information gain between different sub-regions by a hierarchical global prior. This proposed hierarchical global prior contains context information of different scales which varies among different sub-regions. This pyramid pooling module is added at the end of final layer feature map.

This method presents four different pyramid scales to separate feature map into distinct sub-regions to represent pooled area of different locations. These four level pyramid pooling kernels extend over whole, half of, and small portions of the image and are used as global prior in the network. PSPNet uses pretrained ResNet with dilated network to extract feature map. Then 4-level pyramid pooling is applied on the feature map to extract global context prior. These global priors are then concatenated with the original feature map which is followed by a convolutional layer to produce final prediction map. 


\begin{table}[htp]
	\centering
	\caption{Per-class results on HRUD testing set for experiment 1.}\label{table:michael-test-exp1-perfor-table}
	\begin{tabular}{l c c |c}
	\hline
		Method & Building & Road & mIoU\\\hline
		\hline
		ENet\cite{paszke2016enet} & 95.37 & 94.78 & 95.08 \\
		DeepLabV3+\cite{chen2018encoder-deeplabv3+} & 79.9 & 75.0 & 77.45\\ 
		PSPNet\cite{zhao2017pyramid-pspnet} & \bftab 99.87 & \bftab 99.86 & \bftab 99.87\\ \hline
	\end{tabular}
\end{table}


\begin{table*}[htp]
	\centering
	\caption{Per-class results on HRUD testing set for experiment 2.}\label{table:michael-test-exp2-perfor-table}
	\begin{tabular}{l c c c c |c}
	\hline
		Method & \begin{tabular}{@{}c@{}}Building \\  No Damage\end{tabular} & \begin{tabular}{@{}c@{}}Building \\  Medium Damage\end{tabular} & \begin{tabular}{@{}c@{}}Building \\  Major Damage\end{tabular} & \begin{tabular}{@{}c@{}}Building \\  Total Destruction\end{tabular} & mIoU\\\hline
		\hline
		ENet\cite{paszke2016enet} & 62.47 & 42.10 & 34.50 & 88.96 & 57.01\\	
		DeepLabV3+\cite{chen2018encoder-deeplabv3+} & 71.9 & 61.1 & 58.4 & 60.5 & 62.98\\ 
		PSPNet\cite{zhao2017pyramid-pspnet} & \bftab 99.81 & \bftab 99.69 & \bftab 99.69 & \bftab 99.99 & \bftab 99.79\\ \hline
	\end{tabular}
\end{table*}

\begin{table*}[htp]
	\centering
	\caption{Per-class results on HRUD testing set for experiment 3.}\label{table:michael-test-exp3-perfor-table}
	\begin{tabular}{l c c c c c c c c c| c}
	\hline
		Method & Debris & Water &  \begin{tabular}{@{}c@{}}Building Non \\  Total Destruction\end{tabular} & \begin{tabular}{@{}c@{}}Building \\  Total Destruction\end{tabular} & Vehicle & Road & Tree & Pool & Sand & mIoU\\\hline
		\hline
		ENet\cite{paszke2016enet} & 45.97 & 75.84 & 66.16 & 39.52 & 36.74 & 61.19 & 71.64 & 28.47 & 61.77 & 54.15\\
		DeepLabV3+\cite{chen2018encoder-deeplabv3+} & 65.8 & \bftab 85.8 & 84.5 & 57.3 & 51.3 & 73.3 & 75.9 & 55.7 & \bftab 77.4 & 69.67\\ 
		PSPNet\cite{zhao2017pyramid-pspnet} & \bftab 88.76 & 67.98 & \bftab 85.75 & \bftab 80.51 & \bftab 65.83 & \bftab 82.81 & \bftab 94.53 & \bftab 72.61 & 76.04 & \bftab 79.43\\ \hline
	\end{tabular}
\end{table*}

\subsection{DeepLabv3+}
DeepLabv3+ \cite{chen2018encoder-deeplabv3+}, extends DeepLabv3 \cite{chen2017rethinking-deeplabv} by adding a simple yet effective decoder module to refine the segmentation results especially along object boundaries. DeepLabv3 implements several parallel atrous convolution with different rates (called Atrous Spatial Pyramid Pooling, or ASPP), while PSPNet \cite{zhao2017pyramid-pspnet} performs pooling operations at different grid scales. Atrous convolution, also known as Dilated convolution, performs strided convolution on feature maps with different rates. Although rich semantic information is encoded in the last feature map of PSPNet, detailed information related to object boundaries is missing due to the pooling or convolutions with striding operations within the network backbone. This could be alleviated by applying the atrous convolution to extract denser feature maps. Encoder-decoder models \cite{ronneberger2015u-unet, badrinarayanan2017segnet} lend themselves to faster computation (since no features are dilated) in the encoder path and gradually recover sharp object boundaries in the decoder path.

In this structure, one can arbitrarily control the resolution of extracted encoder features by atrous convolution to trade-off precision and runtime, which is not possible with existing encoder-decoder models. Authors further explore the Xception model and apply the depthwise separable convolution to both Atrous Spatial Pyramid Pooling and decoder modules, resulting in a faster and stronger encoder-decoder network.

\begin{figure}

  \begin{subfigure}{0.5\textwidth}
  \centering
    \includegraphics[width=0.8\linewidth]{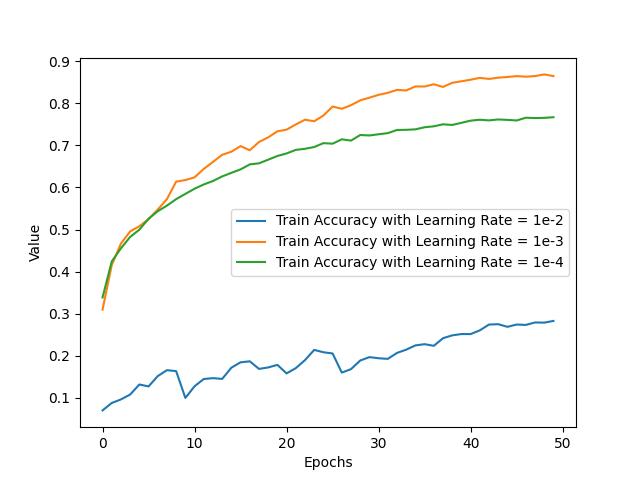}
    \caption{PSPNet} 
    \label{fig:pspnet-hyper-tune-lr}
  \end{subfigure}%
  \\
  \begin{subfigure}{0.5\textwidth}
    \centering
    \includegraphics[width=0.8\linewidth]{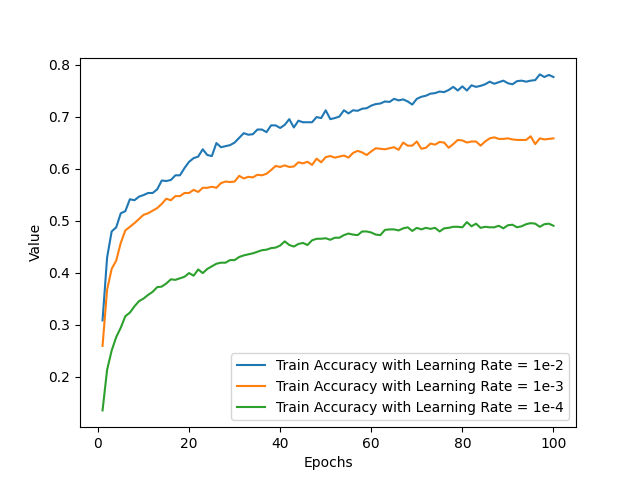}
    \caption{DeepLabV3+} 
    \label{fig:deeplabv3+-hyper-tune-lr}
  \end{subfigure}

\caption{Training Accuracy (mIoU) of (a)PSPNet and (b)DeepLabv3+ with respect to Base Learning Rate.} 
\label{fig:hyper-tune-lr}
\end{figure}

\section{Experiments} \label{experiment}

\subsection{Implementation Details}
Pytorch has been used for implementation of segmentation networks. As hardware we use NVIDIA GeForce RTX 2080 Ti GPU and Intel Core i9 CPU. We implement three segmentation methods, PSPNet \cite{zhao2017pyramid-pspnet}, ENet \cite{paszke2016enet}, and DeepLabv3+ \cite{chen2018encoder-deeplabv3+}, and evaluate their performance on HRUD dataset. For implementing PSPNet, resnet101 has been used as backbone. We use ``poly'' learning rate with base learning rate 0.0001. Momentum, weight decay, power, and weight of the auxiliary loss are set to 0.9, 0.0001, 0.9, and 0.4 respectively. 
For ENet we use 0.0005 and 0.1 for learning rate and learning rate decay respectively. Weight decay is set to 0.0002. Similarly for DeepLabv3+ we use poly learning rate with base learning rate 0.01. We set weight decay to 0.0001 and momentum to 0.9. For augmentation we use random shuffling, scaling, flipping, and random rotation which help models to avoid overfitting.

From different experiments it is proved that larger ``crop size'' and ``batch size'' improve the performance of the models. Due to the limitation of physical memory on GPU card, batch size is set to 2 and crop size is set to 713.



\section{Ablation Study} \label{ablation}
\subsection{Hyperparameter Tuning}
We implement three different segmentation algorithms on HRUD. We vary our base learning rate for poly learning rate scheduler. For the best two methods, DeepLabv3+ and PSPNet, we perform hyperparameter tuning for base learning rate. We vary the base learning rate from 0.01 to 0.0001 for both models. It is found that the best base learning rate for DeepLabv3+ is 0.01, and for PSPNet it is 0.001. The training accuracy (mean IoU) curves with respect to different base learning rates are shown in Figure \ref{fig:hyper-tune-lr}.

\subsection{Distinct Class Accuracy}
Three types of experiments are performed on HRUD. In the first experiment, only buildings and roads are considered for segmentation since from emergency response point of view these two are the most important components in an area. In the second experiment, segmentation is performed on buildings based on their damage levels. For any rescue attempt it is critical to know the damage level of each particular building so that the emergency rescue team can plan their rescue tasks and allocate aids accordingly. Four different building damage levels are considered: no damage, medium damage, major damage, and total destruction. Finally in the third experiment, all types of components are segmented which includes debris, water, building, vehicle, road, tree, pool, and sand.  

\subsubsection{Experiment 1: Building and Road}

To detect only buildings and roads, we merge all types of buildings, annotated separately based on four damage levels, into one single building class. All network models show significant performance on segmentation of buildings and roads. All three methods perform very well on segmenting buildings and roads. Compared to other two experiments ENet \cite{paszke2016enet} shows better performance than DeepLabv3+ \cite{chen2018encoder-deeplabv3+} in this experiment. But PSPNet \cite{zhao2017pyramid-pspnet} performs the best with a near perfect segmentation result with mean IoU of 99.87. Qualitative results can be seen in Figure \ref{fig:vis-compare-models-exp-1} and quantitative results are shown in Table \ref{table:michael-test-exp1-perfor-table}. From the experimental results it is evident that pyramid pooling modules as global context prior are able to segment the buildings and roads more precisely compared to encoder-decoder based method ENet \cite{paszke2016enet} and even DeepLabv3+ \cite{chen2018encoder-deeplabv3+}. Although DeepLabv3+ also implements global context prior in the design of segmentation model, pyramid pooling in PSPNet shows superior performance compared to atrous convolution used in DeepLabv3+.

\subsubsection{Experiment 2: Different Types of Buildings}
We analyze segmentation performances on four different building categories: Building-No-Damage, Building-Medium-Damage, Building-Major-Damage, and Building-Total-Destruction. The network models show that although Building-Total-Destruction and Building-No-Damage can be segmented with very good accuracy, it is very difficult to differentiate between Building-Medium-Damage and Building-Major-Damage. PSPNet \cite{zhao2017pyramid-pspnet} performs the best among all three models in segmenting different types buildings classified based on their respective damage levels. Similar to experiment 1, pyramid pooling module as global context prior is superior to astrous convolution. Specially the segmentation results on class Building-No-Damage and Building-Total-Destruction are almost perfect with mean IoU of 99.81 and 99.99 respectively as shown in Table \ref{table:michael-test-exp2-perfor-table}. Qualitative results are shown in Figure \ref{fig:vis-compare-models-exp-2}. A common aspect of all three performances is that all three networks perform worst in detecting buildings with medium and major damages.

\subsubsection{Experiment 3: All Classes}
In HRUD, in total 8 categories have been annotated. These are building, road, debris, water, pool, tree, sand, and vehicle. As previously mentioned the building category has four different classes: Building-No-Damage, Building-Medium-Damage, Building-Major-Damage, and Building-Total-Destruction. To detect all classes, we merge building with damage levels no damage, medium damage, and major damage into a single class Building. Therefore, including two different building classes (building, building-total-destruction), there are in total 9 classes. 

Evaluations of all three network models on test set presented in Table \ref{table:michael-test-exp3-perfor-table}. Among all these three network models PSPNet \cite{zhao2017pyramid-pspnet} performs the best in test set. These results can be verified from the Figure \ref{fig:vis-compare-models-all-cls} and Figure \ref{fig:vis-compare-models-all-cls-val} which  show that PSPNet \cite{zhao2017pyramid-pspnet} produces predicted images which are closer to the ground truth. Despite good performance on test set, the predicted images produced by DeepLabv3+ \cite{chen2018encoder-deeplabv3+} are not good in terms of edges. PSPNet \cite{zhao2017pyramid-pspnet} produces results with sharp edges and better class prediction.

All the models show a common characteristic regarding their performance on the dataset. The four classes with lowest IoU are pool, vehicle, building-total-destruction, and debris. The lower performance on vehicle is mainly due to the scenarios where vehicles can be found in the middle of debris. As mentioned in subsection \ref{section:dataset-challenges}, if cars are in the middle of debris, they considered as part of debris. These two different appearance of cars bring down the accuracy of the models on detecting cars. The lower IoU of pool is caused by the lower number of training instances. Building-total-destruction is mostly a rubble of debris with a concrete base. Very similar textures and appearances cause the lower performances on segmenting building-total-destruction and debris in all three models.  

\section{Discussion} \label{discussion}
HRUD is a very challenging dataset due to its variable sized classes along with similar textures among classes. Debris makes a great impact on segmentation performances of the evaluated network models. Similar textures of debris, sand, and building with total destruction damage are very difficult to differentiate from both segmentation performances' point of view.

Results from experiment 2 clearly show the difficulty in distinguishing between buildings with medium damage and major damage. Since significant structural damage on the roof is the only differentiating factor between these two damage levels as discussed in \ref{section:data-annotation}, from top view very little information can be obtained about it. This causes poor performances in segmenting these two building damage classes in ENet \cite{paszke2016enet} and DeepLabv3+  \cite{chen2018encoder-deeplabv3+}. Global pyramid pooling module in PSPNet \cite{zhao2017pyramid-pspnet} performs significantly superior than atrous convolution of DeepLabv3+ and encoder-decoder structure based segmentation network ENet. 

From all the results produced by the three models, ENet \cite{paszke2016enet}, PSPNet \cite{zhao2017pyramid-pspnet}, and DeepLabv3+ \cite{chen2018encoder-deeplabv3+}, it is evident that PSPNet \cite{zhao2017pyramid-pspnet} performs best on HRUD dataset. Also from the Figure \ref{fig:pspnet-hyper-tune-lr} and Figure \ref{fig:deeplabv3+-hyper-tune-lr} it is evident that PSPNet \cite{zhao2017pyramid-pspnet} learns the dataset better than DeepLabv3+ \cite{chen2018encoder-deeplabv3+}. Despite lowest performance from ENet \cite{paszke2016enet}, both quantitative and qualitative results are fairly good on HRUD dataset. Performance varies significantly with the variation of base learning rate for both PSPNet \cite{zhao2017pyramid-pspnet}, and DeepLabv3+ \cite{chen2018encoder-deeplabv3+}. Overall PSPNet \cite{zhao2017pyramid-pspnet} shows the best result on the dynamic nature of the HRUD dataset. Pyramid pooling modules in the PSPNet is able to collect global context of all the classes better than atrous convolution in DeepLabv3+. Superior performance of PSPNet than ENet proves that pyramid pooling based segmentation model performs better than encoder-decoder based model on semantic segmentation of HRUD dataset. 

\section{Conclusion} \label{conclusion} 
In this paper, we propose a new high resolution natural disaster dataset named HRUD and evaluate the performance of three popular state-of-art semantic segmentation models on it. We discuss the challenges of semantic segmentation on this dataset along with the reasons of lower performances of the models on certain classes. The network models are investigated thoroughly to achieve the best results out of them for HRUD. We believe our dataset will facilitate future research on natural disaster assessment which will pave the way for less human and economic loss with efficient natural disaster management.
\vspace{12pt}

\section{Acknowledgment}

This work is supported in part by Microsoft. Annotations are performed on the V7 Darwin platform. 

\bibliographystyle{ieeetr}
\bibliography{semantic-segmentation-Michael}

\begin{thebibliography}{10}

\bibitem{noaa}
``Noaa national centers for environmental information (ncei). u.s.
  billion-dollar weather and climate disasters.''
  \url{https://www.ncdc.noaa.gov/billions/events}.
\newblock Accessed: 2020-08-08.

\bibitem{chen2018benchmark}
S.~A. Chen, A.~Escay, C.~Haberland, T.~Schneider, V.~Staneva, and Y.~Choe,
  ``Benchmark dataset for automatic damaged building detection from
  post-hurricane remotely sensed imagery,'' {\em arXiv preprint
  arXiv:1812.05581}, 2018.

\bibitem{gupta2019creating}
R.~Gupta, B.~Goodman, N.~Patel, R.~Hosfelt, S.~Sajeev, E.~Heim, J.~Doshi,
  K.~Lucas, H.~Choset, and M.~Gaston, ``Creating xbd: A dataset for assessing
  building damage from satellite imagery,'' in {\em Proceedings of the IEEE
  Conference on Computer Vision and Pattern Recognition Workshops}, pp.~10--17,
  2019.

\bibitem{lopez2017river}
L.~Lopez-Fuentes, C.~Rossi, and H.~Skinnemoen, ``River segmentation for flood
  monitoring,'' in {\em 2017 IEEE International Conference on Big Data (Big
  Data)}, pp.~3746--3749, IEEE, 2017.

\bibitem{doshi2018satellite}
J.~Doshi, S.~Basu, and G.~Pang, ``From satellite imagery to disaster
  insights,'' {\em arXiv preprint arXiv:1812.07033}, 2018.

\bibitem{rahnemoonfar2018flooded}
M.~Rahnemoonfar, R.~Murphy, M.~V. Miquel, D.~Dobbs, and A.~Adams, ``Flooded
  area detection from uav images based on densely connected recurrent neural
  networks,'' in {\em IGARSS 2018-2018 IEEE International Geoscience and Remote
  Sensing Symposium}, pp.~1788--1791, IEEE, 2018.

\bibitem{rudner2019multi3net}
T.~G. Rudner, M.~Ru{\ss}wurm, J.~Fil, R.~Pelich, B.~Bischke,
  V.~Kopa{\v{c}}kov{\'a}, and P.~Bili{\'n}ski, ``Multi3net: segmenting flooded
  buildings via fusion of multiresolution, multisensor, and multitemporal
  satellite imagery,'' in {\em Proceedings of the AAAI Conference on Artificial
  Intelligence}, vol.~33, pp.~702--709, 2019.

\bibitem{gupta2020rescuenet}
R.~Gupta and M.~Shah, ``Rescuenet: Joint building segmentation and damage
  assessment from satellite imagery,'' {\em arXiv preprint arXiv:2004.07312},
  2020.

\bibitem{gupta2020deep}
A.~Gupta, S.~Watson, and H.~Yin, ``Deep learning-based aerial image
  segmentation with open data for disaster impact assessment,'' {\em arXiv
  preprint arXiv:2006.05575}, 2020.

\bibitem{zhu2020msnet}
X.~Zhu, J.~Liang, and A.~Hauptmann, ``Msnet: A multilevel instance segmentation
  network for natural disaster damage assessment in aerial videos,'' {\em arXiv
  preprint arXiv:2006.16479}, 2020.

\bibitem{ess2009segmentation}
A.~Ess, T.~M{\"u}ller, H.~Grabner, and L.~J. Van~Gool, ``Segmentation-based
  urban traffic scene understanding.,'' in {\em BMVC}, vol.~1, p.~2, Citeseer,
  2009.

\bibitem{geiger2012we}
A.~Geiger, P.~Lenz, and R.~Urtasun, ``Are we ready for autonomous driving? the
  kitti vision benchmark suite,'' in {\em 2012 IEEE Conference on Computer
  Vision and Pattern Recognition}, pp.~3354--3361, IEEE, 2012.

\bibitem{cordts2016cityscapes}
M.~Cordts, M.~Omran, S.~Ramos, T.~Rehfeld, M.~Enzweiler, R.~Benenson,
  U.~Franke, S.~Roth, and B.~Schiele, ``The cityscapes dataset for semantic
  urban scene understanding,'' in {\em Proceedings of the IEEE conference on
  computer vision and pattern recognition}, pp.~3213--3223, 2016.

\bibitem{cao2017realtime}
Z.~Cao, T.~Simon, S.-E. Wei, and Y.~Sheikh, ``Realtime multi-person 2d pose
  estimation using part affinity fields,'' in {\em Proceedings of the IEEE
  Conference on Computer Vision and Pattern Recognition}, pp.~7291--7299, 2017.

\bibitem{yang2016multi}
Z.~Yang and R.~Nevatia, ``A multi-scale cascade fully convolutional network
  face detector,'' in {\em 2016 23rd International Conference on Pattern
  Recognition (ICPR)}, pp.~633--638, IEEE, 2016.

\bibitem{yoon2015learning}
Y.~Yoon, H.-G. Jeon, D.~Yoo, J.-Y. Lee, and I.~So~Kweon, ``Learning a deep
  convolutional network for light-field image super-resolution,'' in {\em
  Proceedings of the IEEE international conference on computer vision
  workshops}, pp.~24--32, 2015.

\bibitem{wan2014deep}
J.~Wan, D.~Wang, S.~C.~H. Hoi, P.~Wu, J.~Zhu, Y.~Zhang, and J.~Li, ``Deep
  learning for content-based image retrieval: A comprehensive study,'' in {\em
  Proceedings of the 22nd ACM international conference on Multimedia},
  pp.~157--166, 2014.

\bibitem{sermanet2013overfeat}
P.~Sermanet, D.~Eigen, X.~Zhang, M.~Mathieu, R.~Fergus, and Y.~LeCun,
  ``Overfeat: Integrated recognition, localization and detection using
  convolutional networks,'' {\em arXiv preprint arXiv:1312.6229}, 2013.

\bibitem{long2015fully-fcn}
J.~Long, E.~Shelhamer, and T.~Darrell, ``Fully convolutional networks for
  semantic segmentation,'' in {\em Proceedings of the IEEE conference on
  computer vision and pattern recognition}, pp.~3431--3440, 2015.

\bibitem{mottaghi_cvpr14-pascal}
R.~Mottaghi, X.~Chen, X.~Liu, N.-G. Cho, S.-W. Lee, S.~Fidler, R.~Urtasun, and
  A.~Yuille, ``The role of context for object detection and semantic
  segmentation in the wild,'' in {\em IEEE Conference on Computer Vision and
  Pattern Recognition (CVPR)}, 2014.

\bibitem{chen2018encoder-deeplabv3+}
L.-C. Chen, Y.~Zhu, G.~Papandreou, F.~Schroff, and H.~Adam, ``Encoder-decoder
  with atrous separable convolution for semantic image segmentation,'' in {\em
  Proceedings of the European conference on computer vision (ECCV)},
  pp.~801--818, 2018.

\bibitem{zhao2017pyramid-pspnet}
H.~Zhao, J.~Shi, X.~Qi, X.~Wang, and J.~Jia, ``Pyramid scene parsing network,''
  in {\em Proceedings of the IEEE conference on computer vision and pattern
  recognition}, pp.~2881--2890, 2017.

\bibitem{paszke2016enet}
A.~Paszke, A.~Chaurasia, S.~Kim, and E.~Culurciello, ``Enet: A deep neural
  network architecture for real-time semantic segmentation,'' {\em arXiv
  preprint arXiv:1606.02147}, 2016.

\bibitem{nguyen2017damage}
D.~T. Nguyen, F.~Ofli, M.~Imran, and P.~Mitra, ``Damage assessment from social
  media imagery data during disasters,'' in {\em Proceedings of the 2017
  IEEE/ACM International Conference on Advances in Social Networks Analysis and
  Mining 2017}, pp.~569--576, 2017.

\bibitem{jegou2017one-tiramisu}
S.~J{\'e}gou, M.~Drozdzal, D.~Vazquez, A.~Romero, and Y.~Bengio, ``The one
  hundred layers tiramisu: Fully convolutional densenets for semantic
  segmentation,'' in {\em Proceedings of the IEEE conference on computer vision
  and pattern recognition workshops}, pp.~11--19, 2017.

\bibitem{isola2017image-pix2pix}
P.~Isola, J.-Y. Zhu, T.~Zhou, and A.~A. Efros, ``Image-to-image translation
  with conditional adversarial networks,'' in {\em Proceedings of the IEEE
  conference on computer vision and pattern recognition}, pp.~1125--1134, 2017.

\bibitem{chen2017rethinking-deeplabv}
L.-C. Chen, G.~Papandreou, F.~Schroff, and H.~Adam, ``Rethinking atrous
  convolution for semantic image segmentation,'' {\em arXiv preprint
  arXiv:1706.05587}, 2017.

\bibitem{chen2017deeplab}
L.-C. Chen, G.~Papandreou, I.~Kokkinos, K.~Murphy, and A.~L. Yuille, ``Deeplab:
  Semantic image segmentation with deep convolutional nets, atrous convolution,
  and fully connected crfs,'' {\em IEEE transactions on pattern analysis and
  machine intelligence}, vol.~40, no.~4, pp.~834--848, 2017.

\bibitem{grauman2005pyramid}
K.~Grauman and T.~Darrell, ``The pyramid match kernel: Discriminative
  classification with sets of image features,'' in {\em Tenth IEEE
  International Conference on Computer Vision (ICCV'05) Volume 1}, vol.~2,
  pp.~1458--1465, IEEE, 2005.

\bibitem{lazebnik2006beyond}
S.~Lazebnik, C.~Schmid, and J.~Ponce, ``Beyond bags of features: Spatial
  pyramid matching for recognizing natural scene categories,'' in {\em 2006
  IEEE Computer Society Conference on Computer Vision and Pattern Recognition
  (CVPR'06)}, vol.~2, pp.~2169--2178, IEEE, 2006.

\bibitem{liu2015parsenet}
W.~Liu, A.~Rabinovich, and A.~C. Berg, ``Parsenet: Looking wider to see
  better,'' {\em arXiv preprint arXiv:1506.04579}, 2015.

\bibitem{newell2016stacked}
A.~Newell, K.~Yang, and J.~Deng, ``Stacked hourglass networks for human pose
  estimation,'' in {\em European conference on computer vision}, pp.~483--499,
  Springer, 2016.

\bibitem{lin2017feature}
T.-Y. Lin, P.~Doll{\'a}r, R.~Girshick, K.~He, B.~Hariharan, and S.~Belongie,
  ``Feature pyramid networks for object detection,'' in {\em Proceedings of the
  IEEE conference on computer vision and pattern recognition}, pp.~2117--2125,
  2017.

\bibitem{shrivastava2016beyond}
A.~Shrivastava, R.~Sukthankar, J.~Malik, and A.~Gupta, ``Beyond skip
  connections: Top-down modulation for object detection,'' {\em arXiv preprint
  arXiv:1612.06851}, 2016.

\bibitem{fu2017dssd}
C.-Y. Fu, W.~Liu, A.~Ranga, A.~Tyagi, and A.~C. Berg, ``Dssd: Deconvolutional
  single shot detector,'' {\em arXiv preprint arXiv:1701.06659}, 2017.

\bibitem{ronneberger2015u-unet}
O.~Ronneberger, P.~Fischer, and T.~Brox, ``U-net: Convolutional networks for
  biomedical image segmentation,'' in {\em International Conference on Medical
  image computing and computer-assisted intervention}, pp.~234--241, Springer,
  2015.

\bibitem{badrinarayanan2017segnet}
V.~Badrinarayanan, A.~Kendall, and R.~Cipolla, ``Segnet: A deep convolutional
  encoder-decoder architecture for image segmentation,'' {\em IEEE transactions
  on pattern analysis and machine intelligence}, vol.~39, no.~12,
  pp.~2481--2495, 2017.

\bibitem{noh2015learning}
H.~Noh, S.~Hong, and B.~Han, ``Learning deconvolution network for semantic
  segmentation,'' in {\em Proceedings of the IEEE international conference on
  computer vision}, pp.~1520--1528, 2015.

\bibitem{lin2017refinenet}
G.~Lin, A.~Milan, C.~Shen, and I.~Reid, ``Refinenet: Multi-path refinement
  networks for high-resolution semantic segmentation,'' in {\em Proceedings of
  the IEEE conference on computer vision and pattern recognition},
  pp.~1925--1934, 2017.

\bibitem{pohlen2017full}
T.~Pohlen, A.~Hermans, M.~Mathias, and B.~Leibe, ``Full-resolution residual
  networks for semantic segmentation in street scenes,'' in {\em Proceedings of
  the IEEE Conference on Computer Vision and Pattern Recognition},
  pp.~4151--4160, 2017.

\bibitem{peng2017large}
C.~Peng, X.~Zhang, G.~Yu, G.~Luo, and J.~Sun, ``Large kernel matters--improve
  semantic segmentation by global convolutional network,'' in {\em Proceedings
  of the IEEE conference on computer vision and pattern recognition},
  pp.~4353--4361, 2017.

\bibitem{amirul2017gated}
M.~Amirul~Islam, M.~Rochan, N.~D. Bruce, and Y.~Wang, ``Gated feedback
  refinement network for dense image labeling,'' in {\em Proceedings of the
  IEEE Conference on Computer Vision and Pattern Recognition}, pp.~3751--3759,
  2017.

\bibitem{wojna2017devil}
Z.~Wojna, J.~R. Uijlings, S.~Guadarrama, N.~Silberman, L.-C. Chen, A.~Fathi,
  and V.~Ferrari, ``The devil is in the decoder.,'' in {\em BMVC}, 2017.

\bibitem{fu2019stacked}
J.~Fu, J.~Liu, Y.~Wang, J.~Zhou, C.~Wang, and H.~Lu, ``Stacked deconvolutional
  network for semantic segmentation,'' {\em IEEE Transactions on Image
  Processing}, 2019.

\bibitem{zhang2018exfuse}
Z.~Zhang, X.~Zhang, C.~Peng, X.~Xue, and J.~Sun, ``Exfuse: Enhancing feature
  fusion for semantic segmentation,'' in {\em Proceedings of the European
  Conference on Computer Vision (ECCV)}, pp.~269--284, 2018.

\bibitem{V7Darwin}
``V7 darwin.'' \url{https://www.v7labs.com/darwin}.
\newblock Accessed: 2020-08-25.

\bibitem{brostow2008segmentation-camvid}
G.~J. Brostow, J.~Shotton, J.~Fauqueur, and R.~Cipolla, ``Segmentation and
  recognition using structure from motion point clouds,'' in {\em European
  conference on computer vision}, pp.~44--57, Springer, 2008.

\bibitem{he2016deep-resnet}
K.~He, X.~Zhang, S.~Ren, and J.~Sun, ``Deep residual learning for image
  recognition,'' in {\em Proceedings of the IEEE conference on computer vision
  and pattern recognition}, pp.~770--778, 2016.

\bibitem{zhou2017scene-ade20k}
B.~Zhou, H.~Zhao, X.~Puig, S.~Fidler, A.~Barriuso, and A.~Torralba, ``Scene
  parsing through ade20k dataset,'' in {\em Proceedings of the IEEE Conference
  on Computer Vision and Pattern Recognition}, 2017.

\bibitem{he2015spatial-sppnet}
K.~He, X.~Zhang, S.~Ren, and J.~Sun, ``Spatial pyramid pooling in deep
  convolutional networks for visual recognition,'' {\em IEEE transactions on
  pattern analysis and machine intelligence}, vol.~37, no.~9, pp.~1904--1916,
  2015.

\end{thebibliography}

\end{document}